\documentclass[conference]{IEEEtran}
\IEEEoverridecommandlockouts
\usepackage{cite}
\usepackage{amsmath,amssymb,amsfonts}
\usepackage{algorithm}
\usepackage{amsthm}
\usepackage[noend]{algpseudocode}
\usepackage{graphicx}
\usepackage{textcomp}
\usepackage{xcolor}

\def\BibTeX{{\rm B\kern-.05em{\sc i\kern-.025em b}\kern-.08em
    T\kern-.1667em\lower.7ex\hbox{E}\kern-.125emX}}
\newtheorem{example}{Example}
\begin{document}

\title{Fine-Tuned Large Language Models for Logical Translation: Reducing Hallucinations with Lang2Logic
}

\author{
\IEEEauthorblockN{Muyu Pan}
\IEEEauthorblockA{\textit{Computer Science and Engineering} \\
\textit{Pennsylvania State University}\\
Pennsylvania State University \\
mfp5696@psu.edu}

\and
\IEEEauthorblockN{Dheeraj Kodakandla}
\IEEEauthorblockA{\textit{Computer Science and Engineering} \\
\textit{Pennsylvania State University}\\
Pennsylvania State University \\
djk6439@psu.edu}
\and

\IEEEauthorblockN{{Mahfuza Farooque}
\IEEEauthorblockA{\textit{Computer Science and Engineering} \\
\textit{Pennsylvania State University}\\
University Park, USA \\
mff5187@psu.edu}
} }

\maketitle

\begin{abstract}
Recent advances in natural language processing (NLP), particularly large language models (LLMs), have motivated the automatic translation of natural language statements into formal logic without human intervention. This enables automated reasoning and facilitates debugging, finding loop invariants, and adhering to specifications in software systems. However, hallucinations-incorrect outputs generated by LLMs are challenging, particularly for logical translation tasks requiring precision. This work introduces a novel framework that inputs English sentences, converts them into logical expressions, and then translates them into Conjunctive Normal Form (CNF) for satisfiability solving. It employs classical NLP techniques with self-defined grammar, symbolic computation libraries, and a fine-tuned language model to reduce hallucinations. In the early experiments, we observed that the fine-tuned model, trained on different grammar settings, could intentionally correct the same types of hallucinations made by the original model. Thus, it provides reliable CNF generation.
 

\end{abstract}


\section{Introduction}
\label{sec:typesetting-summary}

Natural Language Processing (NLP)\cite{NLP} was initially conceptualized by Swiss linguist Ferdinand de Saussure, who introduced the idea that language meaning is created through internal relationships and contrasts. Shared linguistic structures enable communication. In 1950, Alan Turing proposed the concept of a "thinking machine," suggesting that a machine capable of communicating with humans through a teleprinter demonstrates cognitive capability. Contemporary NLP plays a critical role in understanding human language and generating contextually appropriate responses, exemplified by intelligent assistants like Apple's Siri and Amazon's Alexa, which provide personalized assistance and process user requests autonomously. 

Large Language Models (LLMs)\cite{LLM} are sophisticated artificial intelligence models constructed using deep learning methodologies, trained on extensive datasets, and capable of generating human-like textual content. Grounded in transformer architecture, these models are designed to capture complex linguistic nuances and long-range textual dependencies, enabling advanced capabilities such as machine translation, conversational interaction, and content generation. LLMs not only comprehend human languages but also demonstrate applicability across diverse research and industrial domains. OpenAI’s ChatGPT\cite{ChatGPT} serves as a prominent example of LLM technology utilized extensively in daily applications. 

Hallucination\cite{Huang_2025} in language models represents a phenomenon where, based on memorized training data patterns, the model generates outputs containing fabricated, plausible-sounding information when confronted with unseen scenarios. The consequences of hallucinations can range from minor inconsistencies that cause user confusion to critically significant errors in sensitive domains such as language translation, software development, or autonomous systems. Mitigating hallucinations\cite{llmhallucination} in LLMs is paramount for ensuring reliability, safety, and practical applicability, particularly when deploying these models in critical or sensitive contexts. 

To address the challenge of hallucinations, fine-tuned models\cite{ChatGPT} have emerged as an effective solution. These models are pre-trained machine learning models optimized for specialized task domains, demonstrating superior performance compared to generalized models through targeted training on smaller, domain-specific datasets. During the fine-tuning process, model parameters are meticulously adjusted to enhance precision and generalization capabilities. This approach leverages the foundational language understanding acquired during initial large-scale training, subsequently refining the model's focus on specific target tasks. 

Recent works such as LogicLLaMA\cite{yang2023harnessingpowerlargelanguage} and LOGIC-LM\cite{pan2023logiclmempoweringlargelanguage} have pioneered advancements in logical reasoning by fine-tuning LLMs for specialized tasks. LogicLLaMA fine-tunes LLaMA on a dataset of verified NL-FOL pairs to translate natural language to first-order logic (FOL) and mitigate hallucinations using reinforcement learning with human feedback (RLHF). Similarly, LOGIC-LM integrates LLMs with symbolic solvers, converting NL into structured symbolic formulations for deterministic inference while using solver feedback to self-refine and improve accuracy on logical reasoning benchmarks. These studies highlight the importance of fine-tuning and feedback loops in reducing model-generated errors, particularly in logic translation tasks. 

Building on these innovations, this research introduces Lang2Logic, a novel framework designed to bridge the gap between natural language and computational logic. Lang2Logic transforms natural language inputs into Conjunctive Normal Form (CNF), a required format for logical reasoning tasks, facilitating the translation of human-readable statements into machine-processable logical representations. The Lang2Logic framework incorporates symbolic computation libraries and a custom grammar-based parsing approach to ensure accurate transformation from natural language to CNF. 

This structured pipeline enables an end-to-end conversion from unstructured natural language to a canonical Satisfiability (SAT) problem\cite{farooque2024neurodual} representation, making SAT solvers accessible for real-world problems requiring human-readable inputs.

\section{Methodology}

\subsection{Lang2Logic: Translating Natural Language to CNF}
The Lang2Logic framework aims to bridge the gap between natural language understanding and computational logic by converting English text into CNF. This transformation is essential for SAT solvers, as they operate on CNF representations. Lang2Logic consists of three core components: (1) Natural Language to Logical Expression, (2) Logical Expression to CNF Conversion, and (3) CNF Simplification. Each component is critical in ensuring an accurate and efficient transformation from natural language to a machine-readable CNF representation.\\

\subsubsection{Natural Language to Logical Expression}
The first stage utilizes the ChatGPT o1-mini model~\cite{ChatGPT} API to translate natural language input into a logical expression. The prompt for the o1-mini model was carefully designed to:
\begin{itemize}
    \item Check and recognize repeated variables across different clauses,
    \item Output fixed logical expressions in a structured format compatible with downstream processing and
    \item Ensure that the output aligns with the required input format for the \texttt{Sympy} converter~\cite{meurer2017sympy}.
\end{itemize}

Using the ChatGPT API removes the need for hard-coded grammar definitions, simplifying the conversion process. However, the approach has limitations, such as higher costs and potential instability in performance as the input size increases. The NLTK~\cite{bird2009natural} was employed to preprocess the input to mitigate these issues.
Using NLTK's Punkt tokenizer, the input text is split into individual sentences. This ensures that the o1-mini model processes each sentence independently, preventing errors such as sentence merging or skipping during translation. By working sentence-by-sentence, the o1-mini model can produce more accurate logical expressions. Example~\ref{ex:engtoLog} demonstrates the translation of an English sentence into an equivalent propositional logical expression.

\begin{example}
\label{ex:engtoLog}
An English sentence and its logical expression are mentioned below:\\
\textbf{Input Sentence:} \emph{"The circus has a Ferris wheel or a rollercoaster."}\\
\textbf{Logical Expression:} $\text{Or}(P, Q)$, where $P$ represents \emph{"The circus has a ferris wheel"} and $Q$ represents \emph{"The circus has a rollercoaster."}
\end{example}

\subsubsection{Logical Expression to CNF Conversion}
Once the natural language input has been converted to logical expressions, the next stage involves transforming these expressions into CNF. This is accomplished using a Lark parser~\cite{larkparser} and a predefined grammar. Each logical expression is processed as follows:
\begin{enumerate}
    \item \textbf{Parsing:} Each line is passed through the Lark parser to generate a parse tree. The parse tree organizes variables and logical operators into a structured hierarchy.
    \item \textbf{Sympy Conversion:} A custom function reads the parse tree and translates it into a \texttt{Sympy} expression. \texttt{Sympy} then handles the conversion of these expressions into CNF form.
\end{enumerate}



\begin{example}
\label{ex:logtocnf}
Let $P$ be \emph{"The circus does not have a carousel"}, $Q$ be \emph{"The circus has a ferris wheel"}, and $R$ be \emph{"The circus has a rollercoaster"} and let the logical expression be
$\text{And}(\text{Not}(P), \text{Or}(Q, R))$.

The Lark parser generates a structured hierarchy where:
\begin{itemize}
\item $Not(P)$ is identified as a negation operator applied to $P$.
\item $Or(Q, R)$ is identified as a disjunction between $Q$ and $R$.
\item $And(\cdot)$ connects these components into a single logical expression.
  \end{itemize}
\textbf{CNF Output} $\sim P \land (Q \lor R)$.
\end{example}

\subsubsection{CNF Simplification}
The final module simplifies the CNF expressions for computational efficiency. This step leverages SymPy's \texttt{simplify\_logic} method, which reduces redundant clauses and literals while preserving logical equivalence and minimizing complexity. Simplifying the CNF formula ensures the resulting expression is computationally optimal for downstream tasks, such as SAT solving. For instance, an input statement $P \land (\neg Q \lor R) \land (\neg Q \lor R)$ is simplified as $P \land (\neg Q \lor R)$.


The simplification process eliminates redundant clauses, retaining only the unique logical components of the formula. This ensures the CNF is compact while preserving its satisfiability properties, enabling more efficient processing by SAT solvers.\\

\subsubsection{Hallucination Identifying and Removel}
Hallucinations manifest as incorrectly reused variables, misclassified literals, and other translation errors that prevent SAT solvers from operating continuously. The causes of hallucinations in LLMs vary; for instance, these models may fail to accurately translate specific syntactic structures during training. Additionally, non-canonical sentence constructions that deviate from strict word order further complicate the translation into CNF.

In our approach, we combined the NLTK and Sympy libraries to develop a framework for identifying hallucinations. We leverage NLTK for syntactic parsing to ensure that input sentences are accurately decomposed according to rigorously designed grammar. These grammar rules, validated through years of research under careful human supervision, guarantee consistent and correct parsing outcomes. By integrating NLTK with Sympy, our method harnesses precise syntactic analysis and ensures that the subsequent CNF transformation faithfully represents the logical structure of the original sentence.

When a paragraph of natural language text is received, each sentence is treated as an individual clause. The text is split at periods so that each clause is processed independently. Later, these clauses are combined using a logical conjunction operator to form a CNF.

A large-scale, custom-defined context-free grammar (CFG) is used with NLTK to parse each sentence into a parse tree. This parse tree, constructed from the CFG, begins with the root symbol “S” and branches into subcomponents such as IFCLAUSE, ORCLAUSE, and STATEMENT, reflecting the inherent structure of natural language syntax. The tree further includes nodes for determiners (DET), nouns (NOUN), adjectives (ADJ), auxiliary verbs (AUX), and punctuation (PUNC/PUNCT). By converting this tree into a logical expression using predefined rules, we obtain an expression composed of Sympy operators. Sympy then leverages its internal functions to convert each logical expression into CNF.




Hallucinations tend to occur in specific patterns. Once a hallucination is identified by the detection mechanism, the affected segment is immediately reported to the fine-tuned model, establishing a feedback loop. In this loop, the model learns from its errors—including the specific error types—and subsequently generates a correct output for future predictions. Specifically, the model’s output is compared with that of the detection mechanism. If a hallucination is detected, the original input, the hallucinated output, and the corrected output are collected and fed back to the model for fine-tuning. The model then uses the same input to produce an updated output.

\begin{algorithm}
\caption{Lang2Logic Algorithm}
\label{alg:HallucinationCleanNLP}
\begin{algorithmic}[1]
\State \textbf{Input:} Natural language text
\State \textbf{Output:} Final simplified CNF logical expression
\State \textbf{Initialize Grammar, Transformer, and NLP Environment:} 
\State \hspace{0.5cm} Define the Lark grammar for function-call syntax.
\State \hspace{0.5cm} Implement the \texttt{LogicTransformer} to convert tokens into Sympy symbols and logical function calls (e.g., \texttt{And}, \texttt{Or}, \texttt{Not}, etc.).
\State $simplifiedExpr \gets \texttt{simplifyLogic()}$ 

\State $sentences \gets \texttt{sentTokenize(text)}$ \hfill{\text{// Tokenization.}}
\State LangToLogic(\textit{sentences}) \hfill{\text{// o1-mini model conversion.}}

\ForAll{lines in the model response}
    \State $parseTree \gets \texttt{parser.parse(lineStripped)}$ 
    \State $expr \gets \texttt{transformer.transform(parseTree)}$
    \State $cnfExpr \gets \texttt{toCnf()}$ \hfill{\text{// Convert to CNF.}}
    \State \texttt{cnf\_exprs.append(cnf\_expr)}  
\EndFor

\State $simplifiedCnf \gets \texttt{simplifyCnfExpression()}$ 
\State \Return simplifiedCnf
\end{algorithmic}
\end{algorithm}

\section{Result}

\begin{figure}[h!]
    \centering
    \includegraphics[scale=0.06]{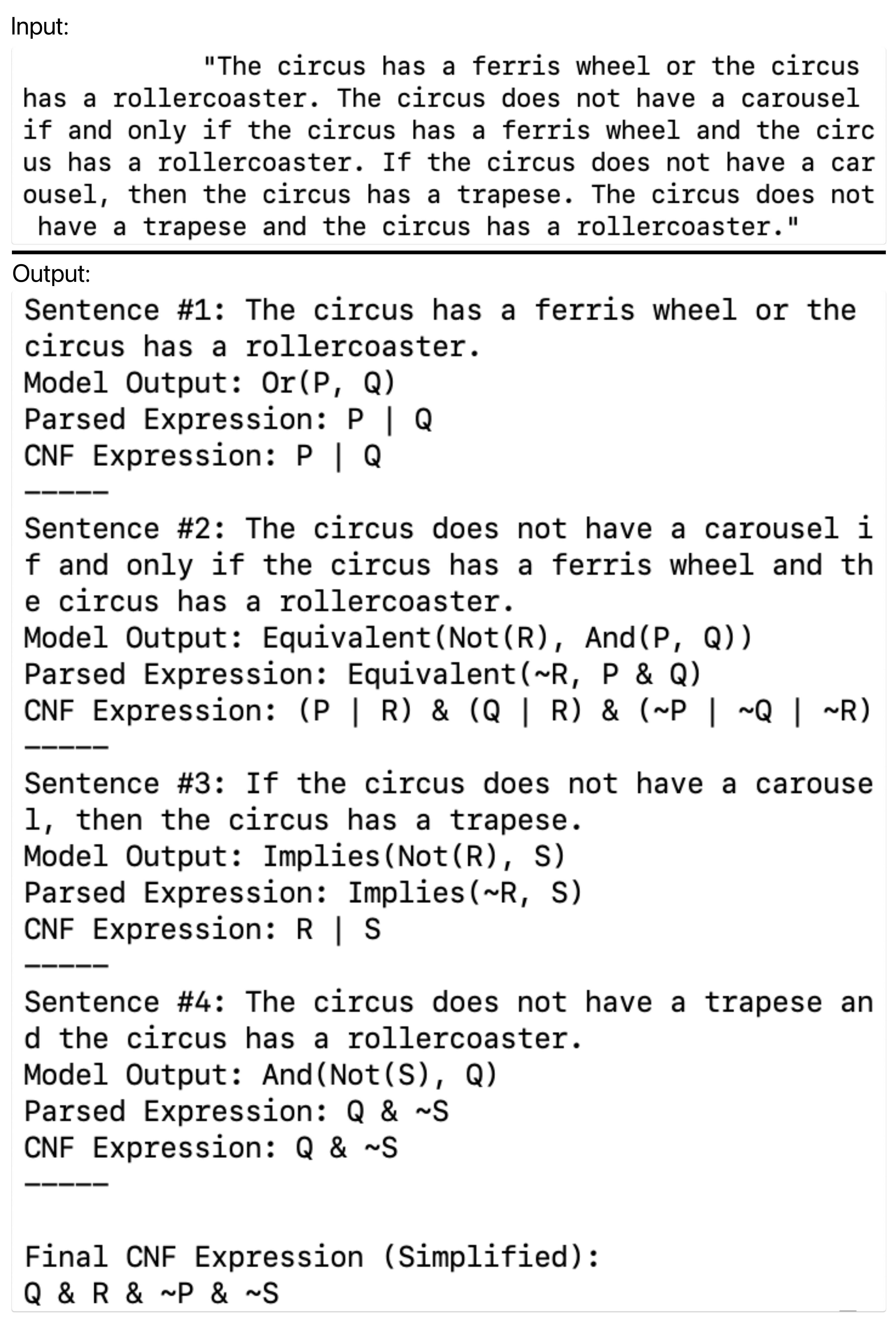} 
    \caption{Lang2Logic test case with four clauses and four variables}
    \label{fig:rl-agent-diagram}
\end{figure}

Lang2Logic digests paragraphs into individual sentences, translates each English sentence into a logical expression, and then converts it to CNF form. It also simplifies the final CNF expression for downstream usage. Figure~\ref{fig:rl-agent-diagram} shows the input paragraph at the top. This paragraph is analyzed by Lang2Logic and converted into four clauses with four variables in CNF form. Lastly, Lang2Logic performs simplifications on this CNF form. The final corresponding output is displayed on the bottom portion of the figure.

\begin{figure}
    \centering
    \includegraphics[scale=0.057]{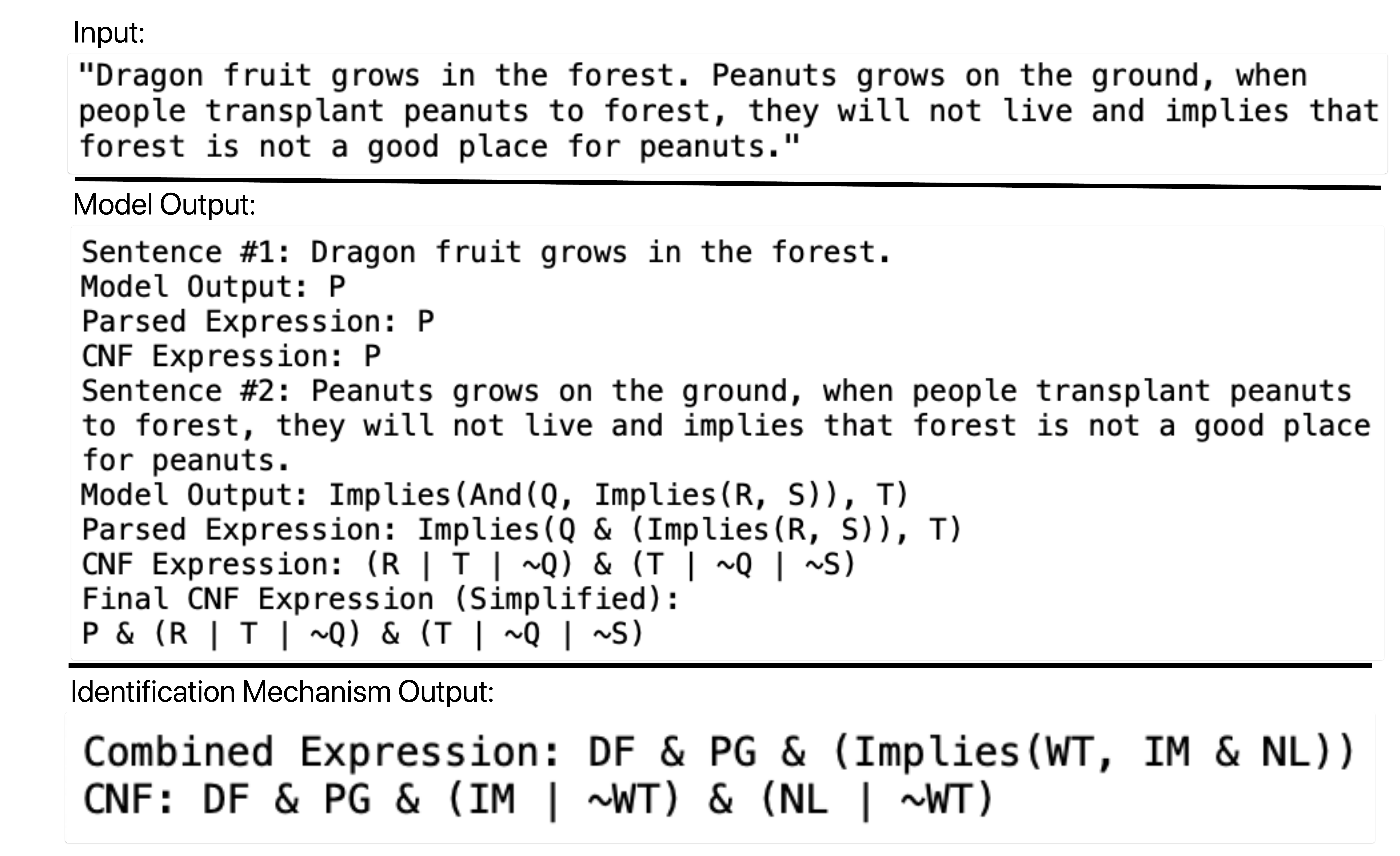} 
    \caption{Hallucination identifying when comparing output from an untuned model with identification mechanism output}
    \label{fig:rlh-agent-diagram}
\end{figure}

Hallucination removal involves identifying and creating a feedback loop to the fine-tuned model, which uses the hallucinated output as an example of the model and corrects the hallucinated component. After feeding back the hallucinated output to the model for fine-tuning, improvement was observed at 100\% on the same hallucination type of inputs. Figure~\ref{fig:rlh-agent-diagram} showcased how the hallucinated output was identified by the identification mechanism output. The untuned model treated ``implies" as a logical implies operator instead of its lexical meaning and used extra ``implies" in the output. But the identification mechanism corrected it.

\section{Conclusion}

In this work, we introduced Lang2Logic, a novel framework designed to bridge the gap between natural language understanding and computational logic. By converting natural language inputs into CNF, Lang2Logic enables more accessible and efficient integration of human-readable statements into machine-processable logical representations. This framework leverages fine-tuned LLMs in conjunction with symbolic computation libraries, ensuring precise and reliable logical translation.
The unique feedback loop integrated into this workflow facilitates the identification and correction of hallucinations, further enhancing the model’s accuracy and robustness. By employing a multi-step pipeline—starting with the translation of natural language to logical expressions, followed by conversion to CNF and simplification—Lang2Logic achieves both computational efficiency and high-quality results suitable for complex logical reasoning tasks.
Overall, this work provides a powerful tool for transforming unstructured text into structured logical formats, offering significant potential for automating and improving logical reasoning tasks across various domains. With its ability to minimize errors and optimize performance, Lang2Logic represents a crucial step forward in enhancing the capabilities of 
LLMs for formal logic applications.

\bibliographystyle{IEEEtran}
\footnotesize
\bibliography{Reference}

\end{document}